\documentclass[10pt,twocolumn,letterpaper]{article}

\usepackage{cvpr}              
\usepackage{amsmath}
\usepackage{amssymb}
\usepackage{amsfonts}
\usepackage{booktabs}
\usepackage{bm}
\usepackage{float}
\usepackage{graphicx}
\usepackage{makecell}
\usepackage{xcolor}
\usepackage{rotating}
\usepackage{textcomp}
\usepackage{arydshln}
\usepackage{xspace}
\usepackage{wasysym}

\definecolor{cvprblue}{rgb}{0.21,0.49,0.74}
\usepackage[pagebackref,breaklinks,colorlinks,allcolors=cvprblue]{hyperref}

\def\paperID{8} 
\def\confName{CVPR}
\def\confYear{2026}

\newcommand{\ours}{{BePo}\xspace}

\title{\ours: Dual Representation for 3D Occupancy Prediction}

\author{
Yunxiao Shi$^{1}$~~
Hong Cai$^{1}$~~
Jisoo Jeong$^{1}$~~
Yinhao Zhu$^{1}$~~
Shizhong Han$^{1}$~~
Amin Ansari$^{2}$~~
Fatih Porikli$^{1}$~~
\smallskip
\\
$^{1}$Qualcomm AI Research\thanks{Qualcomm AI Research is an initiative of Qualcomm Technologies, Inc}\quad $^{2}$Qualcomm Technologies, Inc.
\\
\smallskip
{\tt\small\{yunxshi, hongcai, jisojeon, yinhaoz, shizhan, amina, fporikli\}@qti.qualcomm.com}
}

\begin{document}
\maketitle
\begin{abstract}
3D occupancy infers fine-grained 3D geometry and semantics which is critical for autonomous driving. Most existing approaches carry high compute costs, requiring dense 3D feature volume and cross-attention to effectively aggregate information. More efficient methods adopt Bird's Eye View (BEV) or sparse points as scene representation leading to much reduced runtime. However, BEV struggles with small objects that often have very limited feature representation especially after being projected to the ground plane. Sparse points on the other and, can model objects of various sizes in 3D space, but is inefficient at capturing flat surfaces or large objects. To address these shortcomings, we present \ours, which features a dual representation of BEV and sparse points. The 3D information learned in the sparse points branch is shared with the BEV stream via cross-attention, which injects learning signals of difficult objects on the BEV plane. The outputs of both branches are then fused to generate the final 3D occupancy predictions. Extensive experiments on a suite of challenging benchmarks including Occ3D-nuScenes, Occ3D-Waymo and Occ-ScanNet demonstrate the superiority of our proposed \ours. In addition, \ours carries low inference cost even when compared to latest efficient methods.


\end{abstract}    
\section{Introduction}
\label{sec:intro}

3D occupancy prediction has served as a central task for autonomous driving perception~\cite{wang2022detr3d,liu2023sparsebev,shi2023ega,shi2024decotr,yasarla2023mamo,yasarla2024futuredepth}. Specifically, the task of 3D occupancy prediction aims to infer fine-grained 3D geometry and semantics from camera images, providing critical scene information with a level of fine granularity that is critical for downstream tasks such as motion planning.
\begin{figure}[t!]
    \centering
    \includegraphics[width=0.9\linewidth]{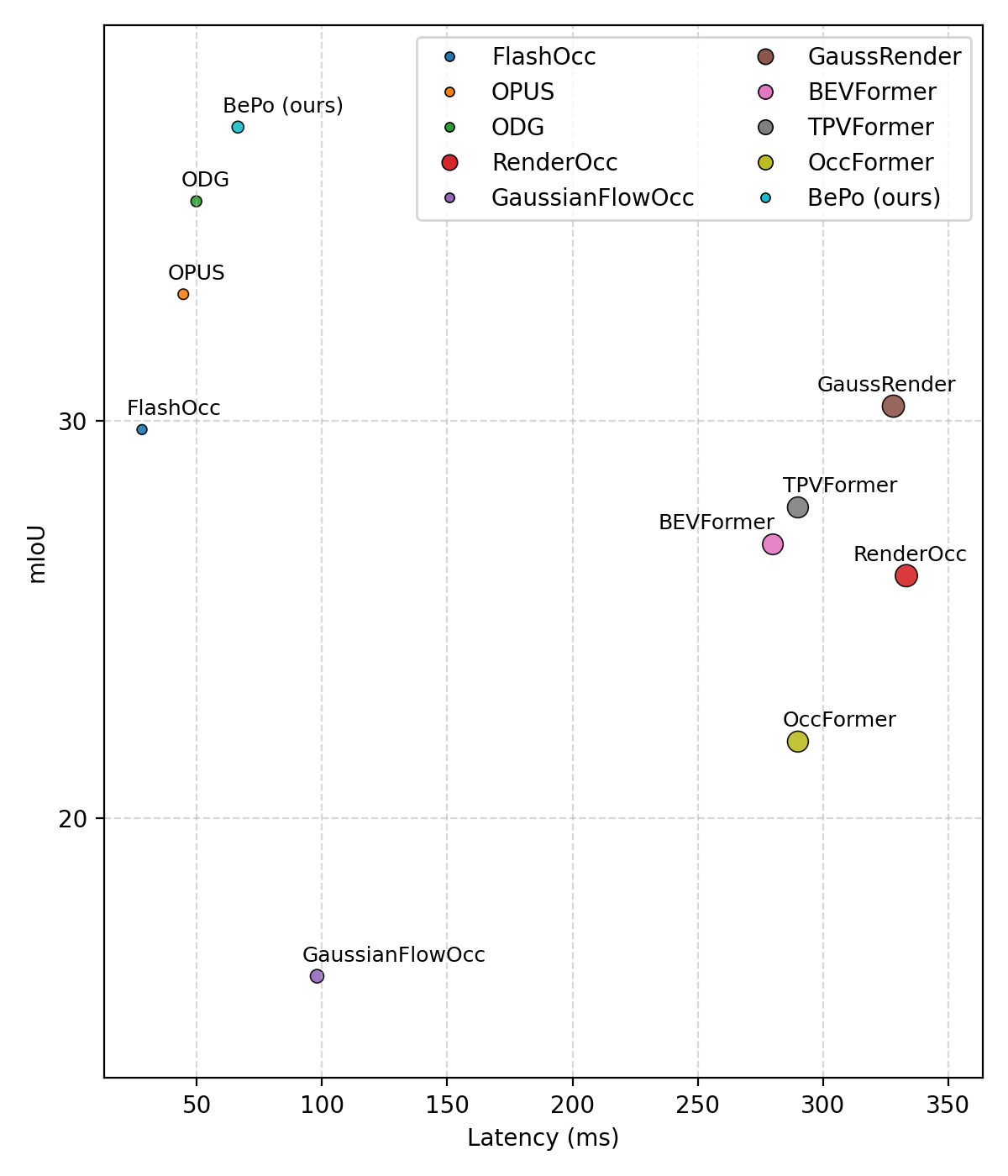}
    \caption{\footnotesize Accuracy (mIoU on Occ3D-nuScenes~\cite{tian2024occ3d, caesar2020nuscenes}) vs. inference latency (ms) measured on a single NVIDIA A100 GPU. ~\ours outperforms previous methods while maintaining competitive inference latency.}
    \label{fig:profile}
\end{figure}

Most existing solutions are computationally expensive, as they adopt a dense 3D feature volume and rely on costly cross-attention to perform 2D-to-3D transformation and feature aggregation~\cite{wei2023surroundocc,zhang2023occformer,pan2024renderocc}. 3D convolutions often follows to process such feature volume incurring significant memory footprint and latency, making it challenging to deploy these models on resource constrained platforms such as autonomous vehicles. To mitigate the high compute cost, BEV representation~\cite{harley2023simple,yu2023flashocc} is advocated by collapsing a 3D volume along the $z$-axis as an alternative and demonstrated much improved inference runtime. Another line of research explores sparse representations, learning the entire 3D scene of a set of points~\cite{wang2024opus, liu2024fully} and achieved competitive performance. Despite the encouraging results, BEV and sparse 3D points as scene representations still suffer from their respective shortcomings. For instance, it is challenging to robustly capture little objects in BEV, as they could have limited 3D representation to begin with due to small size, and further exacerbated when projected onto the BEV plane. Meanwhile, it is not sensible to use sparse 3D points to model simple structures such as flat surfaces, since doing so would require a large number of points where a simple BEV plane can already handle well.

In light of these observations, we propose a new approach, named \ours, which combines the advantages of BEV and sparse 3D point representations. We advocate a dual-branch design in \ours, where one branch first adopts efficient view transform to BEV followed by efficient operators such as 2D convolutions for processing, and the other leverages sparse 3D points with a coarse-to-fine learning scheme. To enable information flow between the two branches, we utilize cross-attention to transfer knowledge from features learned in the points branch to enrich the BEV features. Such learned 3D information from the sparse points can effectively inject more learning signals especially for small objects that have very limited feature representation on the BEV plane.

We note that although two different representations are utilized in our proposed \ours, both feature efficient designs and as a result, \ours still maintains high efficiency. As shown in Figure~\ref{fig:profile}, \ours achieves competitive latency even when compared to the latest efficient approaches. At the same time, \ours sets new State-of-the-Art (SotA) 3D occupancy prediction performance. Our main contributions are summarized as follows:
\begin{itemize}
    \item We propose a novel dual representation, termed \ours, which combines the strengths of both Bird's Eye View (BEV) and 3D sparse points for high-quality 3D occupancy prediction. 
    \item A dual-branch design is advocated where features learned in the sparse points branch are effectively transferred to the BEV branch via cross-attention, which injects learning signals that is lost of objects during BEV projection.  
    \item Extensive experiments on a suite of challenging benchmarks including Occ3D-nuScenes~\cite{tian2024occ3d, caesar2020nuscenes}, Occ3D-Waymo~\cite{tian2024occ3d, sun2020scalability} and Occ-ScanNet~\cite{dai2017scannet,yu2024monocular} demonstrate \ours sets new SotA performance.
\end{itemize}

\begin{figure*}[t!]
    \centering
    \includegraphics[width=0.99\textwidth]{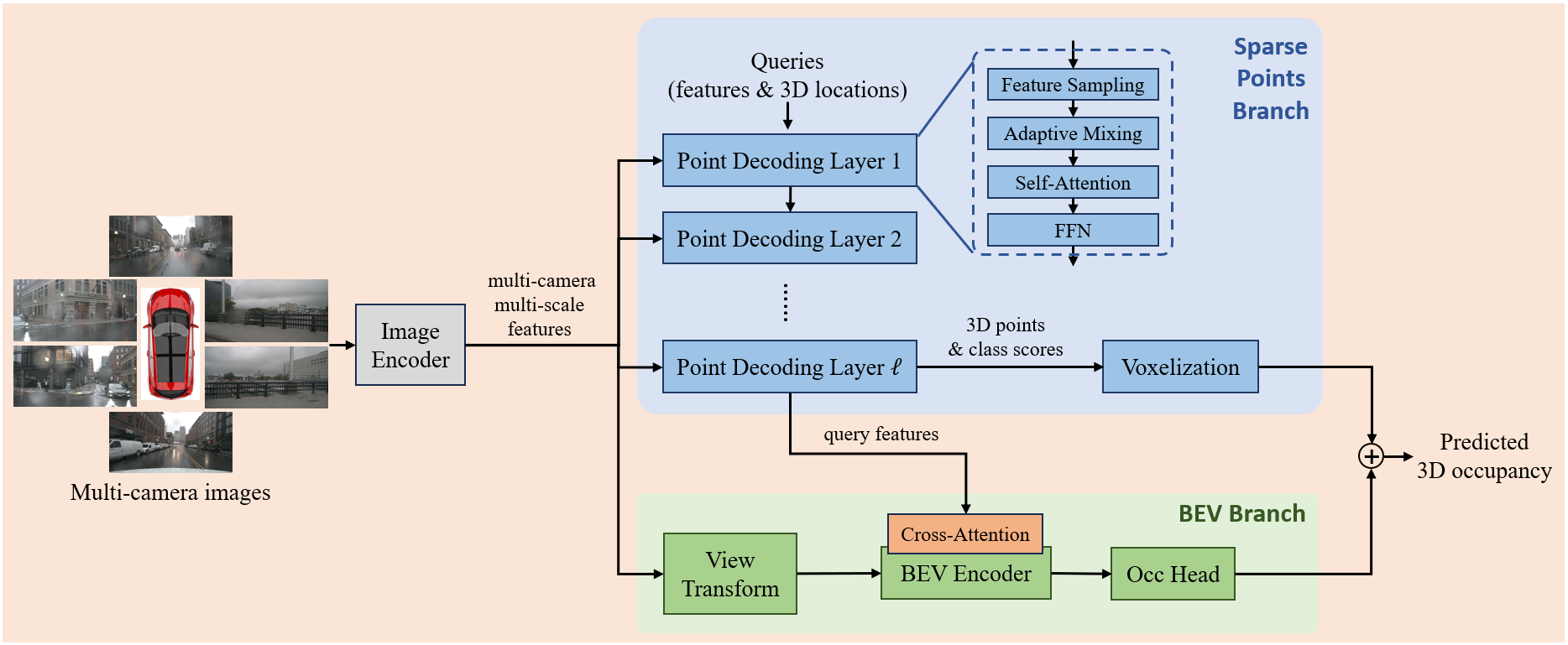}
    \caption{\small Overview of our proposed \ours. First, an image backbone (\eg, ResNet~\cite{he2016deep}) extracts features from the multiple camera images, which are then ingested as input by both the sparse points and BEV branches. Interaction between the features from these two learning streams is enabled via cross-attention. We fuse the volume obtained through voxelization using predicted 3D points locations and class scores from the sparse points branch with the predicted volume from the BEV branch to generate the final predicted 3D occupancy.}
    \label{fig:diagram}
    \vspace{-0pt}
\end{figure*}
\section{Related Work}
\label{sec:related work}

\subsection{3D Occupancy Prediction}

Over the years 3D occupancy prediction has undergone much evolution in terms of methodologies used to better learn 3D scene geometry and semantics. Earlier methods~\cite{cao2022monoscene,li2022bevformer,huang2023tri,wei2023surroundocc,zhang2023occformer} conduct explicit space modeling (\textit{e.g.} voxel, BEV, TPV), then mapped to 3D occupancy and learn from ground-truth data. Due to the intractably high cost of curating occupancy annotations~\cite{wang2023openoccupancy}, ~\cite{zhang2023occnerf,pan2024renderocc,boeder2024occflownet,shi2025h3o} advocates the idea of using 2D labels projected from LiDAR measurements or generated by vision foundation models~\cite{yin2023metric3d,hu2024metric3d} to train 3D occupancy networks, eliminating reliance on direct ground-truth occupancy annotations. Concurrently, several works~\cite{cao2023scenerf,zhang2023occnerf,huang2024selfocc,liu2024let} also explore self-supervised learning based on the photometric consistency between neighboring frames to learn 3D occupancy. Meanwhile, multiple 3D occupancy benchmarks~\cite{behley2019semantickitti,wei2023surroundocc,tian2024occ3d,tong2023scene,openscene2023,wang2023openoccupancy,zhu2024nucraft,yu2024monocular} have been created based on existing datasets~\cite{geiger2012we,geiger2013vision,caesar2020nuscenes,sun2020scalability,dai2017scannet}.

\subsection{Set Prediction for 3D perception}
Direct set prediction with transformers~\cite{vaswani2017attention,dosovitskiy2020image,letourneau2024padre} for perception tasks was first introduced in DETR~\cite{carion2020end} for object detection. Follow-up works like~\cite{wang2022detr3d,liu2022petr,liu2023sparsebev,lin2022sparse4d} further advanced this technique for 3D object detection and demonstrated compelling results. Witnessing such success, ~\cite{tang2024sparseocc,wang2024opus} adapted such set prediction paradigm for 3D occupancy prediction, demonstrating the common pipeline of constructing dense grids followed by classification is not the only way. Inspired by the impressive success of 3D Gaussian Splatting~\cite{kerbl20233d,kerbl2024hierarchical} for scene reconstruction~\cite{charatan2024pixelsplat,chen2024mvsplat,szymanowicz2024splatter}, another line of research extends points to 3D Gaussians for occupancy prediction~\cite{huang2024gaussianformer,huang2025gaussianformer,chambon2025gaussrender,jiang2025gausstr,boeder2025gaussianflowocc,shi2025odg}. In particular, GaussTR~\cite{jiang2025gausstr} and GaussianFlowOcc~\cite{boeder2025gaussianflowocc} employs a single transformer to predict the respective Gaussian properties from sparse queries, albeit can only handle small number of Gaussians limiting model capability. ODG~\cite{shi2025odg} extends to a multi-stage coarse-to-fine prediction paradigm that further improved results.

\section{Proposed Approach: \ours}

We first provide a brief overview on BEV-based and point-based methods for 3D occupancy prediction in Section~\ref{overview}. We then describe the details of the two branches in our proposed BePo in Section~\ref{bev} and Section~\ref{points}. Section~\ref{cross-attn} describes how cross-attention is used between the two branches and Section~\ref{fusion} demonstrates how we fuse the predictions from the two branches. Figure~\ref{fig:diagram} provides an overview of \ours.

\subsection{Overview}\label{overview}
Given an ego-vehicle at time $T$, 3D occupancy prediction takes $N_C$ camera images (with $kN_c$ optional history frames where $k\geq 0$), $\mathbf{I}=\{I_c^t\}_{t=T-k,c=1}^{T,N_C}$ and camera parameters as input and predicts a 3D semantic occupancy volume $\mathbf{O}\in\mathbb{R}^{H\times W\times Z}$,
where $H,W,Z$ denotes the resolution of the volume, $\mathbf{O}_{ijl}\in\{c_1, c_2, ..., c_C\}$ is the number of semantic classes. We can formally describe 3D occupancy prediction as follows:
\begin{equation}
    \mathbf{O}=G(\mathbf{V}),\quad \mathbf{V}=F(\mathbf{I}),
\end{equation}
where $F(\cdot)$ consists of the image backbone that extracts multi-camera features and transforms them to into a scene feature representation $\mathbf{V}$, and $G(\cdot)$ is another neural network that maps $\mathbf{V}$ to occupancy predictions. A typical choice for $\mathbf{V}$ is a dense voxel grid~\cite{wei2023surroundocc}, which ignores the scene sparsity and cannot handle different object scales leading to unbalanced and wasteful resource allocation. BEV-based methods~\cite{li2022bevformer,yu2023flashocc} demonstrate by collapsing the 3D volume $\mathbf{V}$ along the $Z$ axis into a 2D ground plane and encoding height information into the channel dimension can achieve competitive performance while significantly improving overall efficiency. Yet, BEV-based methods struggle at small objects since they have very limited representation on the BEV plane.

To tackle such issues, \cite{liu2023sparsebev,wang2024opus} argues that explicitly building a dense representation $\mathbf{V}$ followed by classification necessitates intricate descriptions of the 3D space, which is inherently plagued by the scene sparsity issue and hinders end-to-end learning. Hence OPUS~\cite{wang2024opus} in particular casts 3D occupancy prediction as a direct set prediction task with transformers. Such a paradigm can be described as
\begin{align}\label{eq:overall}
    \min_{\mathbf{P}, \mathbf{C}}D_p(\mathbf{P}, \mathbf{P}_g) + D_c(\mathbf{C}, \mathbf{C}_g),
\end{align}
where $\{\mathbf{P}_g, \mathbf{C}_g\}$ is the ground-truth set for occupied voxels with $|\mathbf{P}_g| = |\mathbf{C_g}| = V_g$ being the number of occupied voxels. Each $p_g\in\mathbf{P}_g$ represents the center coordinate of the its 3D voxel and $c_g\in\mathbf{C}_g$ the semantic label. Correspondingly $\{\mathbf{P},\mathbf{C}\}$ denotes the set predictions. $D_p(\cdot)$ and $D_c(\cdot)$ are certain geometric and semantic distances respectively. It is worth noting that flexible as such approach is, a large of points is still needed to effectively learn everything in the scene. Given the fact a BEV plane is already sufficient for simple structures such as road surfaces, points should ideally only attend to difficult objects in 3D space, which leads to our proposed ~\ours.

\subsection{Efficient BEV Branch}\label{bev}

We first employ an image backbone to extract multi-scale features $\mathcal{F}_{im}\in\mathbb{R}^{C\times H\times W}$ from the multi-camera input images, where $C, H, W$ are channel, width and height respectively. Then the extracted image features undergo view transform $T$ to be projected into the BEV space. Here we choose $T$ to be LSS~\cite{philion2020lift} to perform such transform given its efficiency and simplicity. Afterwards, a BEV encoder $E$ consisting of a stack of convolutional layers and an FPN~\cite{lin2017feature} neck are used to process the BEV features to obtain $\mathcal{F}_{bev}\in\mathbb{R}^{C_b\times H_b\times W_b}$. To generate the final occupancy prediction $O_{bev}$, we use a lightweight occupancy prediction head and map the processed BEV features to 3D occupancy. Our BEV branch can be summarized as follows:
\begin{align}
    \mathcal{F}_{bev} &= T(\mathcal{F}_{im}),\\
    O_{bev} &= OccHead(E(\mathcal{F}_{bev})).
\end{align}

\subsection{Sparse 3D Points Branch}\label{points}

The 3D points branch shares the same image backbone as the BEV branch as illustrated in Figure~\ref{fig:diagram}. Inspired by recent works on query-based 3D perception~\cite{liu2023sparsebev,wang2024opus,jiang2025gausstr,boeder2025gaussianflowocc,shi2025odg}, we randomly initialize a set of learnable queries $\mathbf{Q}$ and their point locations $\mathbf{P}$. These query features and predictions are then fed into a sequence of transformer decoders and iteratively refined correlating with the image features from the shared image backbone. Formally, denote $\mathcal{S}_i = \{\mathbf{Q}_i, \mathbf{P}_i, \mathbf{C}_i\}_{i=0}^{\ell}$ the sets, where $\mathcal{S}_0$ is the initial set and $\mathcal{S}_{i>0}$ are the outputs from the $i$-th decoder stage. $\ell$ is the number of decoder layers. To reduce computation bottleneck, we follow~\cite{wang2024opus} and make each $q_i\in\mathbf{Q}_i$ predict multiple points instead of one, denoted as $M_i$. A coarse-to-fine prediction paradigm such that $M_{i-1}\leq M_i, \{i=1,\dots, \ell\}$ is adopted to facilitate predicting high-level semantics from low-level features.

The details of our transformer decoder layers are analogous to that of ~\cite{liu2023sparsebev,wang2024opus}, which we briefly summarize below. Given a previous query $q_{i-1}\in\mathbf{Q}_{i-1}$ and its point location $p_{i-1}\in\mathbf{P}_{i-1}$, the $i$-th decoder layer takes them as input and aggregates image features through point sampling~\cite{wang2024opus}. $q_i$ is updated through adaptive mixing and self-attention among all queries as in~\cite{liu2023sparsebev}. The outputs of the $i$-th decoding layer are class scores $c_i\in\mathbb{R}^{M_i\times N}$ and point offsets $\Delta p_i\in\mathbb{R}^{M_i\times 3}$. Point location $p_i$ is then update as
\begin{equation}
    p_i = p_{i-1} + \Delta p_i,
\end{equation}
where the deltas are learned based on query features.

\subsection{BEV-Point Cross Attention}\label{cross-attn}

In BePo, we aim to effectively combine the strengths of the BEV branch and the points branch. Therefore it is important to enable information flow between the two branches. To this end, we compute cross-attention~\cite{vaswani2017attention} between the BEV features $\mathcal{F}_{bev}$ and the query features $q_\ell\in\mathbb{R}^{M_i\times C_q}$ from the last decoding stage of the points branch. Specifically, we treat $F_{bev}$ as queries and derive keys and values from $q_\ell$ such that for regions challenging for the BEV plane to handle, it can assimilate information from the point query features which are more 3D aware. Formally, we first reshape $F_{bev}$ into shape $\mathbb{R}^{H_bW_b\times C_b}$ and project $q_\ell$ through a linear projection $\pi: C_q\mapsto C_b$,
\begin{equation}
    \hat{q}_\ell = \pi(q_\ell)\in\mathbb{R}^{M_i\times C_b}
\end{equation}
to match the embedding dimension. Then cross-attention is computed as
\begin{align}
    &Q = \mathcal{F}_{bev}W_q, K = q_\ell W_k, V = q_\ell W_v,\\
    &\text{Attn}(Q, K, V) = \text{softmax}\Big(\frac{QK^{\top}}{\sqrt{C_b}}\Big)\cdot V,
\end{align}
where $W_q, W_k, W_v$ are corresponding weight matrices.

\subsection{Volume Fusion}\label{fusion}

To further facilitate end-to-end learning in our proposed \ours, we fuse the two volume outputs from the two branches. For the volume $O_p$ from the points branch, the predicted points locations $\mathbf{P}$ and corresponding class scores $\mathbf{C}$ are used to get
\begin{equation}
    O_p = \text{Voxelize}(\mathbf{P}, \mathbf{C}),
\end{equation}
Then the final output $O_f$ is obtained through a fusion operator $\tau$
\begin{equation}
    O_f = \tau(O_{bev}, O_p),
\end{equation}
where we choose $\tau$ to be element-wise addition.

\begin{table*}[t!]
  \small
  \centering
  \caption{\small 3D semantic occupancy prediction results on Occ3D-nuScenes validation set~\cite{caesar2020nuscenes}. $*$ indicates self-supervised methods. \textbf{Bold}/\underline{Underline}: Best/second best results.}
  \vspace{-5pt}
  \resizebox{\textwidth}{!}{
  \begin{tabular}{l|c|cccccccccccccccccc}
    \toprule
    \textbf{Method} & \makecell{mIoU} & \makecell{\begin{turn}{90}Others\end{turn}} & \makecell{\begin{turn}{90}Barrier\end{turn}} & \makecell{\begin{turn}{90}Bicycle\end{turn}} & \makecell{\begin{turn}{90}Bus\end{turn}} & \makecell{\begin{turn}{90}Car\end{turn}}& \makecell{\begin{turn}{90}Cons. veh\end{turn}}& \makecell{\begin{turn}{90}Motorcycle\end{turn}} & \makecell{\begin{turn}{90}Pedestrian\end{turn}} & \makecell{\begin{turn}{90}Traffic cone\end{turn}} & \makecell{\begin{turn}{90}Trailer\end{turn}} & \makecell{\begin{turn}{90}Truck\end{turn}} & \makecell{\begin{turn}{90}Dri. sur\end{turn}} & \makecell{\begin{turn}{90}other flat\end{turn}}& \makecell{\begin{turn}{90}Sidewalk\end{turn}} & \makecell{\begin{turn}{90}Terrain\end{turn}} & \makecell{\begin{turn}{90}Manmade \end{turn}} & \makecell{\begin{turn}{90}Vegetation\end{turn}} & \makecell{RayIoU}\\
    \midrule
    BEVFormer~\cite{li2022bevformer} & 23.67 &5.03 &38.79 &9.98 &34.41 &41.09 &13.24 & 16.50 & 18.15 & 17.83 &18.66 &27.70 &48.95 &27.73 &29.08 &25.38 &15.41 &14.46 & 32.4\\
    OccFormer~\cite{zhang2023occformer} & 21.93 &5.94 &30.29 &12.32 &34.40 &39.17 &14.44 &16.45 &17.22 &9.27 &13.90 &26.36 &50.99 &30.96 &34.66 &22.73 &6.76 &6.97 & -\\
    RenderOcc~\cite{pan2024renderocc} & 26.11 & 4.84 & 31.72 & 10.72 & 27.67 & 26.45 &13.87 &18.2 &17.67 & 17.84 &21.19 &23.25 &63.2 & 36.42 & \textbf{46.21} &44.26 &19.58 & 20.72 & 19.5\\
    GaussRender~\cite{chambon2025gaussrender} &30.38	&8.87	&40.98	&\underline{23.25}	&43.76	&46.37	&19.49	&\underline{25.2} &\underline{23.96}	&\textbf{19.08}	&25.56	&33.65	&58.37	&33.28	&36.41	&33.21	&22.76	&22.19 & 37.5\\
    GaussTR*~\cite{jiang2025gausstr}  &12.27 &-	&6.5	&8.54	&21.77	&24.27	&6.26	&15.48	&7.94	&1.86	&6.1	&17.16	&36.98	&-	&17.21	&7.16	&21.18	&9.99 & -\\
    FlashOcc~\cite{yu2023flashocc} & 29.79 & 4.87& 35.55  & 9.12 &34.62 &  41.89 & 15.45& 13.01& 14.99& 13.58& 25.63 &29.88 & 75.51 & 33.93 & 44.29& \underline{49.28} & 34.44 & 30.38 &-\\
    OPUS~\cite{wang2024opus} &33.20 &10.72 &\underline{39.82} &21.27 &39.76 &45.25 &23.41 &21.80 &17.81 &19.26 &27.48 &33.20 &71.61 &37.12 &45.13 &43.59 &33.80& \underline{33.18} & 38.4\\
    GaussianFlowOcc*~\cite{boeder2025gaussianflowocc}  &16.02	&-	&7.23	&9.33	&17.55	&17.94	&4.5	&9.32	&8.51	&10.66	&2.00	&11.80	&63.89	&-	&31.11	&35.12	&14.64	&12.59 & 16.47\\
    ODG~\cite{shi2025odg} & 35.54 & \underline{13.69} & 38.97 & 23.02 & \textbf{46.75} & \underline{49.33} &\textbf{25.79} &23.63 &20.73 &18.54 &\underline{30.01} & \textbf{35.61} &\underline{76.84} &\underline{39.33} &45.01 & 46.78 &\textbf{37.45}& 32.24 & \underline{39.2} \\
    \midrule
    \ours & \textbf{37.27} &\textbf{14.11} &\textbf{42.37} &\textbf{25.11} &\underline{44.89} &\textbf{50.04} &\underline{25.22} &\textbf{27.66} &\textbf{25.91} & \underline{19.01} &\textbf{32.64} &\underline{35.13} &\textbf{78.97} &\textbf{40.12} &\underline{45.71} &\textbf{54.27} &\underline{37.33} &\textbf{35.15} & \textbf{40.1}\\
    \bottomrule
  \end{tabular}}
  \vspace{-0pt}
  \label{tab:nuscenes}
\end{table*}

\subsection{Loss Functions}

We impose supervisions on the different outputs of proposed \ours. Cross-entropy loss and Lovasz-softmax~\cite{berman2018lovasz} loss are used to supervise the final volume $O_f$
\begin{equation}
    \mathcal{L}_{vol} = \mathcal{L}_{ce}(O_f, O_g) + \mathcal{L}_{lovasz}(O_f, O_g),
\end{equation}
where $O_g$ is the ground-truth 3D semantic occupancy annotations. In addition, we also supervise the predicted points locations and class scores from all decoding stages of the points branch. Chamfer distance~\cite{fan2017point} is used to supervise predicted the points locations $\mathbf{P}$ to encourage the distributions aligns with ground-truths
\begin{equation}
    \mathcal{L}_{loc} = \sum_{i=0}^\ell \text{CD}(\mathbf{P}_i, \mathbf{P}_g).
\end{equation}
For the class scores $\mathbf{C}$, as pointed by~\cite{wang2024opus}, a direct comparison is not valid due to misalignment in correspondence to different location. Hence we adopt the scheme that assigns each predicted point the class of its nearest neighbor as~\cite{wang2024opus} to get the updated ground-truth $\hat{\mathbf{C}}$. Then focal loss~\cite{ross2017focal} is used to supervised the predicted $\mathbf{C}$
\begin{equation}
    \mathcal{L}_{cls} = \text{FocalLoss}(\mathbf{C}, \hat{\mathbf{C}}).
\end{equation}
Therefore the final loss is formulated as
\begin{equation}
    \mathcal{L} = \mathcal{L}_{vol} + \alpha(\mathcal{L}_{loc} + \mathcal{L}_{cls}),
\end{equation}
where $\alpha$ is the weight that balances the loss terms.
\section{Experiments}
\label{sec:experiments}

We evaluate \ours on Occ3D-nuScenes~\cite{caesar2020nuscenes,tian2024occ3d} and Occ3D-Waymo~\cite{sun2020scalability,tian2024occ3d} benchmarks, and compare with latest state-of-the-art methods. To further demonstrate the robustness and generalizability of \ours, we perform evaluation on Occ-ScanNet~\cite{yu2024monocular} as well which features an indoor setup. Extensive ablation studies on the different design choices of \ours are also conducted.

\begin{figure*}[t!]
    \begin{center}$
    \centering
    \begin{tabular}{c}
    \includegraphics[width=0.98\textwidth]{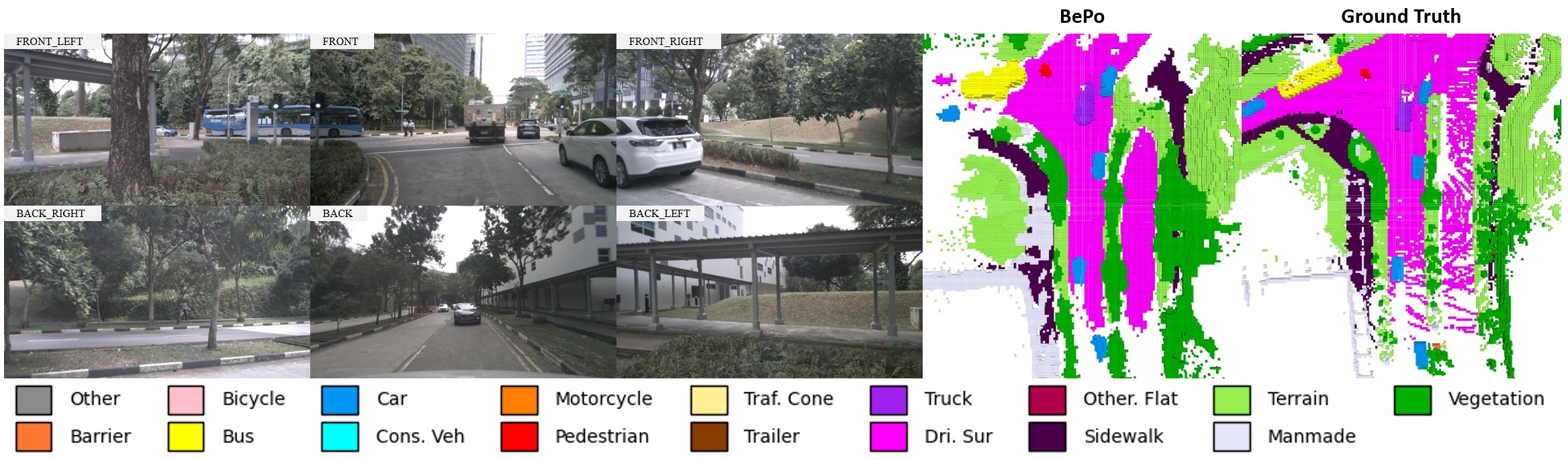}
    \end{tabular}$
    \end{center}
    \vspace{-5pt}
    \caption{\small Example qualitative 3D semantic occupancy prediction of \ours on Occ3D-nuScenes validation set. Cons. Veh stands for ``Construction Vehicle'' and Dri. Sur stands for ``Drivable Surface''. Both prediction and ground-truth are visualized under BEV. Best viewed in color and zoomed in.}
    \label{fig:demo-res}
    \vspace{-0pt}
\end{figure*}

\subsection{Datasets and Setup}

\begin{figure*}[t!]
    \centering
    \includegraphics[width=0.99\linewidth]{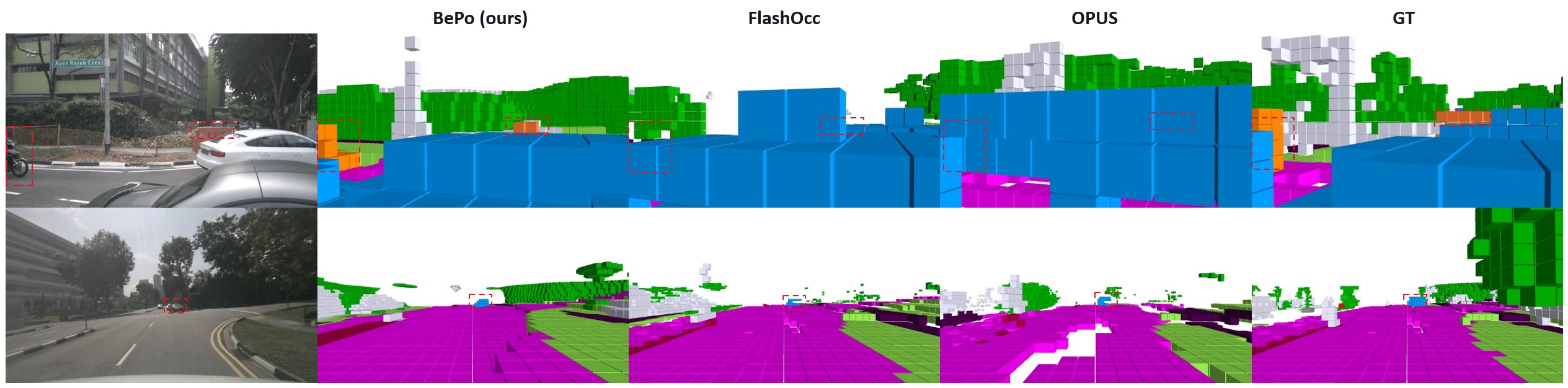}
    \caption{\small Example qualitative comparison of ~\ours with~\cite{yu2023flashocc} and~\cite{wang2024opus} on Occ3D-nuScenes validation set. We see that \ours is able to capture small objects, \eg, motorcycle at frame edge, car at far distance, that are represented by only a very limited number of voxels while both other methods failed.}
    \label{fig:demo-compare}
\end{figure*}

\noindent\textbf{Occ3D-nuScenes}: nuScenes consists of 1,000 scenes captured by a synchronized camera array of 6 cameras. The dataset is split into 700 scenes for training, 150 scenes for validation and 150 scenes for testing. Occ3D-nuScenes boostraps nuScenes and annotates 3D semantic occupancy ground-truth consisting of 17 classes. The voxel grid range is $[-40\text{m}, -40\text{m}, -1\text{m}, 40\text{m}, 40\text{m}, 5.4\text{m}]$ along the $X,Y$ and $Z$ axis. The grid resolution is $200\times 200\times 16$ with a voxel size of $0.4$m. The raw image resolution is $900\times 1600$.

\noindent\textbf{Occ3D-Waymo}: Occ3D-Waymo is curated based on the Waymo Open Dataset~\cite{sun2020scalability}. It features 5 cameras and consists of 798 training scenes and 202 validation scenes. The 3D semantic occupancy ground-truth on Occ3D-Waymo has 15 classes with 1 class being \textit{General Object (GO)}. The voxel range is $[-40m, 40m]$ for the $X$ and $Y$ axis and $[-1m, 5.4m]$ for the $Z$ axis. The voxel size is $0.4$m with a voxel grid resolution of $200\times 200\times 16$. For the front, front-left and front-right cameras, the raw image resolution is $1280\times 1920$. For the side-left and side-right cameras, the raw image resolution is $886\times 1920$.

\noindent\textbf{Occ-ScanNet}: Occ-ScanNet is built on top of the ScanNet~\cite{dai2017scannet} dataset. It provides scenes represented in $60\times 60\times 36$ voxel grids with a voxel size of $0.08m$, leading to a $X,Y,Z$ range of $[4.8m, 4.8m, 2.88m]$. The dataset is labeled with 12 semantic classes where 11 of which are for valid semantics and 1 for free space. We follow~\cite{yu2024monocular} and use a train-validation split of 45,755 frames and 19,764 frames, respectively. The image resolution used for training is $480\times 640$.

\subsection{Evaluation Metrics}

To evaluate 3D semantic occupancy prediction, we compute the mean Intersection over Union (mIoU) of occupied voxels, averaging over all semantic classes
\begin{equation}\label{eq:iou}
    \text{mIoU} = \frac{1}{C}\sum_{i=1}^C\frac{TP_i}{TP_i + FP_i + FN_i},\notag
\end{equation}
where $TP, FP, FN$ denotes the number of true positive, false positive and false negative predictions. We also evaluate on the more recent RayIoU metric proposed in ~\cite{tang2024sparseocc} 
\begin{equation}
    \text{RayIoU} = \frac{\sum_{r\in\mathcal{R}}|P_r\cap G_r|}{\sum_{r\in\mathcal{R}}{|P_r\cup G_r|}},
\end{equation}
where $P,G$ are the set of occupied voxels in prediction and ground-truth respectively. $\mathcal{R}$ is the set of all emulated LiDAR rays, and $P_r, G_r$ are the sets of occupied voxels intersected by ray $r$ in prediction and ground-truth respectively.



\subsection{Implementation Details}

\begin{table*}[h]
  \small
  \centering
  \caption{\small 3D semantic occupancy results on Occ3D-Waymo validation set~\cite{caesar2020nuscenes}. GO stands for ``General Object''. Traf. light stands for ``Traffic light'' and Cons. cone stands for ``Construction cone''. \textbf{Bold}/\underline{Underline}: Best/second best results.}
  \resizebox{\textwidth}{!}{
  \begin{tabular}{l|c|cccccccccccccccc}
    \toprule
    \textbf{Method} & \makecell{mIoU} & \makecell{\begin{turn}{90}GO\end{turn}} & \makecell{\begin{turn}{90}Vehicle\end{turn}} & \makecell{\begin{turn}{90}Bicyclist\end{turn}} & \makecell{\begin{turn}{90}Pedestrian\end{turn}} & \makecell{\begin{turn}{90}Sign\end{turn}}& \makecell{\begin{turn}{90}Traf. light\end{turn}}& \makecell{\begin{turn}{90}Pole\end{turn}} & \makecell{\begin{turn}{90}Cons. cone\end{turn}} & \makecell{\begin{turn}{90}Bicycle\end{turn}} & \makecell{\begin{turn}{90}Motorcycle\end{turn}} & \makecell{\begin{turn}{90}Building\end{turn}} & \makecell{\begin{turn}{90}Vegetation\end{turn}} & \makecell{\begin{turn}{90}Treetrunk\end{turn}}& \makecell{\begin{turn}{90}Road\end{turn}} & \makecell{\begin{turn}{90}Sidewalk\end{turn}} & \makecell{RayIoU}\\
    \midrule
    BEVDet~\cite{huang2021bevdet} & 9.88 &0.13 &13.06 &2.17 & 10.15 &7.80 &5.85 &4.62 &0.94 &1.49 &0.00 &7.27 &10.06 &2.35 &48.15 &34.12 &-\\
    BEVFormer~\cite{li2022bevformer} & 16.76 & 3.48 &17.18 &13.87 &5.9 &13.84 &2.7 &9.82 &12.2 &13.99 &0.00 &13.38 &11.66 &6.73 &74.97 & 51.61 &- \\
    TPVFormer~\cite{huang2023tri} & 16.76 &3.89 &17.86&12.03 &5.67 &13.64 &8.49 &8.90 &9.95 &14.79 &0.32 &13.82 &11.44 &5.8 &73.3 &51.49 &-\\ 
    CTF-Occ~\cite{tian2024occ3d} & 18.73 & \underline{6.26} & 28.09 &14.66 &8.22 & 15.44 & \underline{10.53} & 11.78 & \textbf{13.62} & \underline{16.45} & \underline{0.65} &18.63 &17.3 & \underline{8.29} &67.99 &42.98 &-\\
    OPUS~\cite{wang2024opus} & 19.00 & 4.66 &27.07 & 19.39 &6.53 & \textbf{18.66} &6.41 & 11.44 &10.40 &12.90 &0.00 & 18.73 & \underline{18.11} &7.46 &72.86 &50.31 &24.7\\
    ODG~\cite{shi2025odg} & \underline{21.35} & 5.09 &	\underline{31.34} & \underline{22.4} &	\underline{19.06}  & 15.24 &	6.09 &	\underline{12.51} &	12.77 &	13.59 &	0.00 	&\textbf{21.49} 	&17.89 & 	\textbf{8.37} &	\underline{78.19} &	\textbf{56.28} & \underline{25.9}\\
    \midrule
    \ours & \textbf{23.71} & \textbf{7.79} & \textbf{33.12} & \textbf{25.41} & \textbf{26.18} & \underline{17.78} & \textbf{12.91} & \textbf{15.71} &\underline{13.12} & \textbf{19.21} & \textbf{0.70} & \underline{21.32} &\textbf{20.79} & 8.11 & \textbf{79.54} & \underline{54.07} & \textbf{26.7}\\
    \bottomrule
  \end{tabular}}
  \label{tab:waymo}
\end{table*}

We use ResNet-50~\cite{he2016deep} as the image backbone with pretrained weights from BEVDet~\cite{huang2021bevdet} to extract multi-camera image features for both Occ3D benchmarks. For Occ-ScanNet, following~\cite{yu2024monocular}, we utilize a pretrained EfficientNet-B7~\cite{tan2019efficientnet} backbone. On Occ3D-nuScenes, we resize input images to $256\times 704$. On Occ3D-Waymo, all input images are resized and padded to $640\times 960$. For the points branch, we set the number of queries $Q=600$ and the number of decoding layers $\ell=6$. The number of points sampled for each query is set to 4. The loss weight $\alpha$ is set to $0.1$. We use AdamW~\cite{loshchilov2017decoupled} as the optimizer with weight decay of $0.01$. We train ~\ours with an initial learning rate of $2e-4$ with a warm-up schedule of the first 100 iterations and decays with CosineAnnealing~\cite{loshchilov2016sgdr}. All experiments are conducted on 8 NVIDIA A100 GPUs with a per-GPU batch size of 8 on Occ3D-nuScenes and 4 on Occ3D-Waymo, and trained for 24 epochs. On Occ-ScanNet, following~\cite{yu2024monocular} we train with a global batch size of 8 but for 20 epochs. All inference latencies are measured on a single NVIDIA A100 GPU.


\begin{figure*}[h]
    \begin{center}$
    \centering
    \begin{tabular}{c}
    \includegraphics[width=0.96\textwidth]{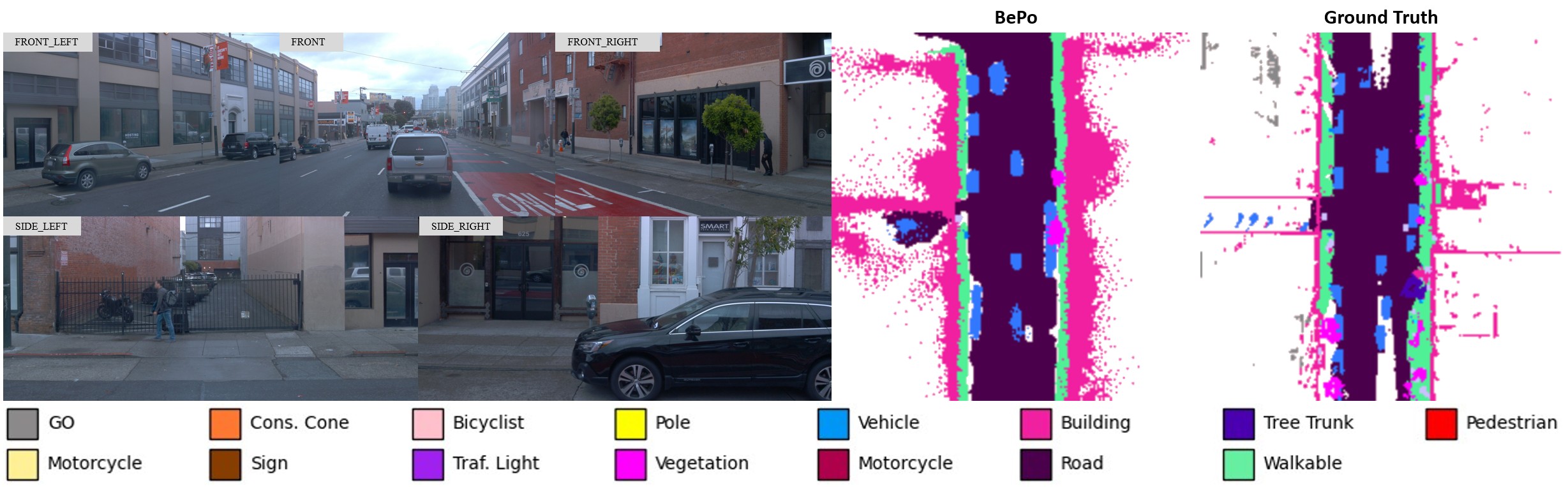}
    \end{tabular}$
    \end{center}
    \vspace{-5pt}
    \caption{\small Example qualitative 3D semantic occupancy prediction of \ours on Occ3D-Waymo validation set. Both prediction and ground-truth are visualized under BEV. Best viewed in color and zoomed in.}
    \label{fig:demo-res-waymo}
    \vspace{-0pt}
\end{figure*}

\subsection{3D Occupancy Prediction Results}
\vspace{0pt}

\textbf{Occ3D-nuScenes.}
Table~\ref{tab:nuscenes} summarizes the 3D semantic occupancy prediction results on Occ3D-nuScenes. \ours sets a new state-of-the-art performance with a mIoU of 37.27 and a RayIoU of 40.1, outperforming existing SotA methods by a significant margin while still maintaining high inference speed.  And if we take a closer look at the per-class IoUs, one can see \ours consistency delivers the best accuracy in small objects such as \textit{Motorcycle}, \textit{Pedestrian}, \textit{Bicycle}, \textit{Others} (general objects), and also flat surfaces such as \textit{Drivable Surface}, validating the effectiveness of our dual representation design.

\begin{table*}[h]
  \footnotesize
  \centering
  \caption{\footnotesize 3D occupancy prediction results on Occ-ScanNet validation set~\cite{yu2024monocular}. \textbf{Bold}/\underline{Underline}: Best/second best results.}
  \resizebox{0.82\textwidth}{!}{
  \begin{tabular}{l|c|c|cccccccccccc}
    \toprule
    \textbf{Method} & \makecell{IoU} & \makecell{mIoU} & \makecell{\scriptsize \begin{turn}{90}Ceiling\end{turn}} & \makecell{\scriptsize \begin{turn}{90}Floor\end{turn}} & \makecell{\scriptsize \begin{turn}{90}Wall\end{turn}} & \makecell{\scriptsize \begin{turn}{90}Window\end{turn}} 
    & \makecell{\scriptsize \begin{turn}{90}Chair\end{turn}}
    & \makecell{\scriptsize \begin{turn}{90}Bed\end{turn}}
    & \makecell{\scriptsize \begin{turn}{90}Sofa\end{turn}} & \makecell{\scriptsize \begin{turn}{90}Table\end{turn}} & \makecell{\scriptsize \begin{turn}{90}TVs\end{turn}} 
    & \makecell{\scriptsize \begin{turn}{90}Furniture\end{turn}} 
    & \makecell{\scriptsize \begin{turn}{90}Objects\end{turn}} \\
    \midrule
    MonoScene~\cite{cao2022monoscene} & 41.60 & 24.62 & 15.17 &\underline{44.71} & \underline{22.41} &12.55 & 26.11 &27.03 &35.91 &28.32 &6.57 &32.16 & 19.84\\
    ISO~\cite{yu2024monocular} & \underline{42.16} & \underline{28.71} & \underline{19.88} & 41.88 & 22.37 &\underline{16.98} &\underline{29.09} &\underline{42.43} & \underline{42.00} & \underline{29.60} & \underline{10.62} & \underline{36.36} &\underline{24.61}\\
    \midrule
    Ours & \textbf{52.73} & \textbf{44.91} & \textbf{41.32} & \textbf{50.29} & \textbf{41.83} & \textbf{31.81} & \textbf{40.37} & \textbf{54.65} & \textbf{60.71} & \textbf{43.76} & \textbf{34.27} & \textbf{53.33} & \textbf{41.72}\\
    \bottomrule
  \end{tabular}}
  \vspace{-2pt}
  \label{tab:scannet}
\end{table*}

Figure~\ref{fig:demo-res} provides a qualitative example of the 3D occupancy prediction by our proposed \ours. Overall, we see that \ours produces clear, accurate 3D geometry of the scene. Specifically, \ours is able to capture objects that appear small in the camera images, \eg, pedestrians (voxels in red). It is also able to capture objects at long distances, such as the car in the back camera (represented by the blue voxels). Such capabilities are critical for safe autonomous driving. 

To provide further intuition that \ours carries the strengths of both BEV and sparse points, we provide qualitative comparison with ~\cite{yu2023flashocc} and ~\cite{wang2024opus} in Figure~\ref{fig:demo-compare}, two representative approaches that utilize BEV and points representations, respectively. One can see that \ours is able to capture small objects that have very limited voxel representation. For instance, for the motorcycle at the frame edge in the top image, \ours correctly predicts its 3D semantic occupancy where both other methods completely missed. In the second example (second row) which has a vehicle at great distance to the back, \ours is able to give complete occupancy predictions of the vehicle, while both other methods are only able to capture parts. This further validates the dual representation of our proposed \ours effectively improves model's capability at difficult cases.

\vspace{5pt}
\noindent\textbf{Occ3D-Waymo.}
Table~\ref{tab:waymo} shows the 3D semantic occupancy prediction results on the Occ3D-Waymo validation dataset in terms of mIoU and RayIoU. Occ3D-Waymo is a more challenging benchmark as there is very little overlap between the cameras and itself is not a commonly used benchmark for evaluating vision-only 3D semantic occupancy prediction. Nonetheless, \ours sets a new SotA performance with a mIoU of 23.71 and a RayIoU of 26.7, significantly outperforming all previous methods that have been evaluated on Occ3D-Waymo. Table~\ref{tab:waymo} further shows that \ours delivers the best accuracy in the majority of classes on Occ3D-Waymo, which include difficult yet safety-critical categories such as \textit{General Object (GO)}, \textit{Bicyclist}, \textit{Pedestrian}, \textit{Traffic Light (Traf. light)} and \textit{Motorcyclist}. Visual example of \ours prediction is shown in Figure~\ref{fig:demo-res-waymo}. One can see the \ours is able to capture all the vehicles in the scene nicely.

\begin{figure}[h]
    \begin{center}$
    \centering
    \begin{tabular}{c}
    \includegraphics[width=0.47\textwidth]{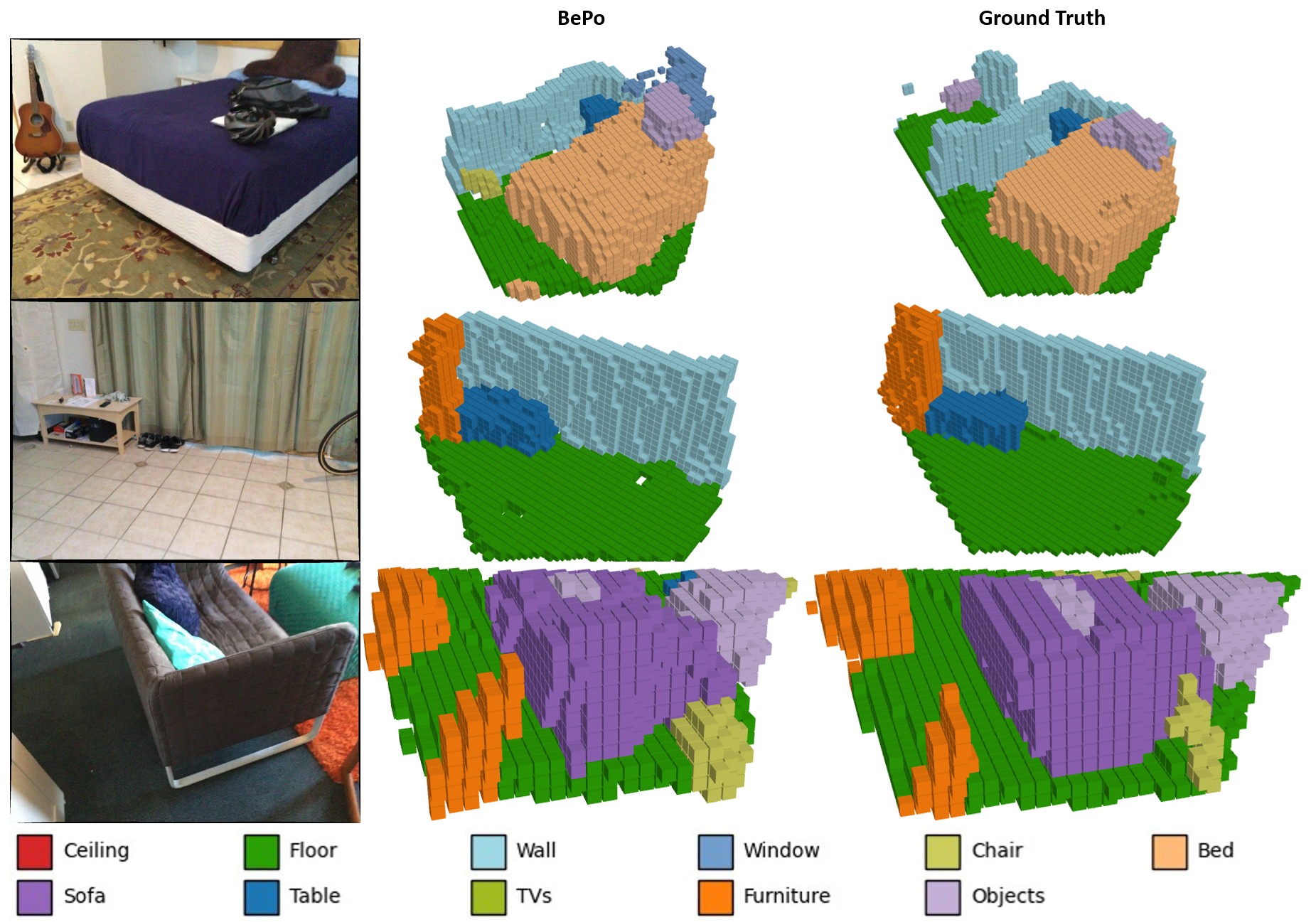}
    \end{tabular}$
    \end{center}
    \vspace{-5pt}
    \caption{\small Example qualitative 3D semantic occupancy prediction of \ours on OccScanNet validation set. Best viewed in color and zoomed in.}
    \label{fig:demo-res-scannet}
    \vspace{-0pt}
\end{figure}

\noindent\textbf{Occ-ScanNet.} To further demonstrate the robustness and generalizability of ~\ours, we conduct experiments under an indoor setup on Occ-ScanNet~\cite{yu2024monocular}. The results are summarized in Table~\ref{tab:scannet}. One can see that ~\ours outperforms previous methods in all objects categories by a large margin in terms of both mIoU and class-agnostic IoU, which validates the effectiveness of our design. Figure~\ref{fig:demo-res-scannet} shows a few example visualizations of \ours on the validation split of Occ-ScanNet. One can see \ours consistently delivers high-quality predictions. We note that our experiment on Occ-ScanNet demonstrates that \ours has the potential to used for indoor applications, such as perception for AR/VR smart glasses.

\subsection{Computation Efficiency Analysis}

\begin{table*}[h]
  \small
  \centering
  \caption{\small Efficiency measurements conducted on NVIDIA A100 GPU with PyTorch FP32. mIoU and RayIoU is on Occ3D-nuScenes~\cite{tian2024occ3d}.}
  \begin{tabular}{p{3cm}p{2.5cm}p{2cm}p{2cm}p{2cm}p{2cm}}
    \toprule
    \textbf{Method} & Backbone & Latency & GPU Memory &mIoU &RayIoU\\
    \midrule
    BEVFormer~\cite{li2022bevformer} & Res101-DCN & 280ms & 4.5G & 27.83 & 32.4\\
    TPVFormer~\cite{huang2023tri} & Res101-DCN & 290ms & 5.1G & 27.83 & -\\
    OccFormer~\cite{zhang2023occformer} & Res101-DCN & 290ms & 9.4G & 21.93 & -\\
    RenderOcc~\cite{pan2024renderocc} & Swin-base  & 420ms & 7.4G & 26.11 &19.5\\
    GaussRender~\cite{chambon2025gaussrender} & Res101-DCN & 328ms & 17.6G & 30.38 & 37.5 \\
    FlashOcc~\cite{yu2023flashocc} & Res-50 & 28.1ms & 2.1G & 29.79 & -\\
    OPUS~\cite{wang2024opus} & Res-50 & 44.6ms & 3.9G & 33.20 & 38.4\\
    GaussianFlowOcc~\cite{boeder2025gaussianflowocc} & Res-50 & 98.0ms & - & 17.08 & 16.47\\
    ODG~\cite{shi2025odg} & Res-50 & 49.8ms & 4.5G & 35.54 & 39.2\\
    \midrule
    \ours (ours) & Res-50 & 66.7ms & 4.1G & \textbf{37.27} & \textbf{40.1}\\
    \bottomrule
  \end{tabular}
  \vspace{-0pt}
  \label{tab:efficiency}
\end{table*}

We conduct analysis on runtime efficiency in Table~\ref{tab:efficiency}. When evaluated on a single NVIDIA A100 GPU, our proposed \ours runs very efficiently with a latency of 66ms and a memory consumption of only 4.1GB.  FlashOcc has the fastest inference speed but underperforms \ours by a large margin. OPUS~\cite{wang2024opus} has the second best runtime but falls short of \ours in both mIoU and RayIoU. ODG~\cite{shi2025odg} also demonstrated fast runtime but came at a cost of higher memory consumption. This analysis demonstrates that \ours strikes a good balance between accuracy and runtime efficiency, delivering good performance on both fronts.

\subsection{Ablation Study}
In this section, we conduct ablation studies to analyze the effects of different parts of our proposed \ours.

\subsubsection{Number of Queries in Sparse Points Branch} 
The number of queries used in \ours's sparse points branch has a significant impact on the final prediction performance and the overall computation bottleneck. Here we conduct experiments with different numbers of queries to observe how it affects accuracy and runtime efficiency. 

The results are shown Table~\ref{tab:query}. We can see that reducing the number of queries results in visible improvements in frames-per-second (FPS) at inference time while also maintaining good prediction performance in terms of both mIoU and RayIoU, only experiencing minor drops. This further demonstrates the effectiveness and robustness of \ours's dual representation.

\begin{table}[h]
  \small
  \centering
  \caption{\small Different numbers of queries in the sparse points branch.}
  \begin{tabular}{lcccc}
    \toprule
    \textbf{Method} & Number of queries & mIoU & RayIoU & FPS\\
    \midrule
     \ours (ours) & \makecell{200\\400\\600} & \makecell{34.17\\36.39\\ 37.27} & \makecell{39.1\\39.8\\40.1} &\makecell{16.0\\15.6\\15.1}\\ 
    \bottomrule
  \end{tabular}
  \vspace{0pt}
  \label{tab:query}
\end{table}

\subsubsection{Effect of Cross-Attention} 

We study the effect of cross attention between the BEV branch and sparse points branch in this part. The results are summarized in Table~\ref{tab:cross-attn}. One can see that when removing cross-attention between the BEV features and query features from the sparse points branch leads to a drop both in mIoU and RayIoU for 3D occupancy prediction. This demonstrates that the cross-attention effectively injects learning signals from 3D features learned by sparse 3D points to enrich the BEV features.  

\begin{table}[h]
  \small
  \centering
  \caption{\small Ablation study on the effect of cross attention between branches.}
  \begin{tabular}{lccc}
    \toprule
    \textbf{Method} & Cross attention & mIoU &RayIoU \\
    \midrule
     \ours (ours) & \makecell{\\ \checkmark} & \makecell{35.39 \\ 37.27} & \makecell{39.5\\40.1}\\ 
    \bottomrule
  \end{tabular}
  \label{tab:cross-attn}
\end{table}

\section{Conclusion}
\label{sec:conclusion}

In this paper, we present \ours, a novel approach for 3D occupancy prediction. \ours advocates a dual representation based on BEV and sparse 3D points, where the BEV branch only has efficient view transform followed by fast 2D operators, and the sparse points branch offers a coarse-to-fine prediction scheme to model difficult scene objects. Cross-attention is computed between the BEV features and the sparse query features to enable information flow. Extensive evaluation on a suite of challenging benchmarks demonstrate that \ours acheives new state-of-the-art results, while maintaining low inference cost.

For future work, we intend to extend \ours to explicitly model motion of dynamic objects which has the potential to further improve 3D occupancy prediction. We also plan to explore using 3D Gaussians as our sparse branch to establish a more object-centric representation, while also leverage the fast rendering of 3D Gaussian splatting to inject more learning signals to complement 3D learning.

{
    \small
    \bibliographystyle{ieeenat_fullname}
    \bibliography{main}
}


\end{document}